\documentclass[10pt,twocolumn,letterpaper]{article}

\usepackage{cvpr}              

\usepackage{graphicx}
\usepackage{amsmath}
\usepackage{amssymb}
\usepackage{booktabs}
\usepackage{lipsum}

\newcommand\blfootnote[1]{%
  \begingroup
  \renewcommand\thefootnote{}\footnote{#1}%
  \addtocounter{footnote}{-1}%
  \endgroup
}

\newcommand{\bblue}[1]{\textcolor{blue}{\textbf{#1}}}

\usepackage[pagebackref,breaklinks,colorlinks]{hyperref}

\usepackage[capitalize]{cleveref}
\crefname{section}{Sec.}{Secs.}
\Crefname{section}{Section}{Sections}
\Crefname{table}{Table}{Tables}
\crefname{table}{Tab.}{Tabs.}

\begin{document}

\title{Monocular Human Digitization via Implicit Re-projection Networks}

\author{Min-Gyu Park, Ju-Mi Kang, Je Woo Kim, and Ju Hong Yoon\\
Korea Electronics Technology Institute (KETI)\\
{\tt\small \{mpark, yypeip, jwkim, jhyoon\}@keti.re.kr}
}
\maketitle

\begin{abstract}
We present an approach to generating 3D human models from images. The key to our framework is that we predict double-sided orthographic depth maps and color images from a single perspective projected image. Our framework consists of three networks. The first network predicts normal maps to recover geometric details such as wrinkles in the clothes and facial regions. The second network predicts shade-removed images for the front and back views by utilizing the predicted normal maps. The last multi-headed network takes both normal maps and shade-free images and predicts depth maps while selectively fusing photometric and geometric information through multi-headed attention gates. Experimental results demonstrate that our method shows visually plausible results and competitive performance in terms of various evaluation metrics over state-of-the-art methods.
\end{abstract}

\section{Introduction}
\label{sec:intro}

\blfootnote{This work was supported by Institute of Information \& communications Technology Planning \& Evaluation (IITP) grant funded by the Korea government(MSIT) (No. 2022-0-00566. The development of object media technology based on multiple video sources) and by Korea Institute for Advancement of Technology(KIAT) grant funded by the Korea Government(MOTIE)(P146500035, The development of interactive metaverse concert solutions via neural human modeling), and by the Technology development Program(S3113187) funded by the Ministry of SMEs and Startups.}

Image-based human digitization has long been an actively researched topic, and it is one of the crucial technologies to the upcoming metaverse era since the reconstructed humans can mirror people in the real world to a virtual space. Commercial products that employ conventional techniques such as multi-view stereo matching and bundle adjustment became famous for generating virtual characters~\cite{Collet_2015_TOG}. Meanwhile, recent advances in deep learning have opened a new direction for human digitization research, \ie a single image-based 3D human model recovery, which allows people to generate their avatars by capturing a single image from their smartphones.

\begin{figure}[t]
\centering
\includegraphics[width=1.0\columnwidth]{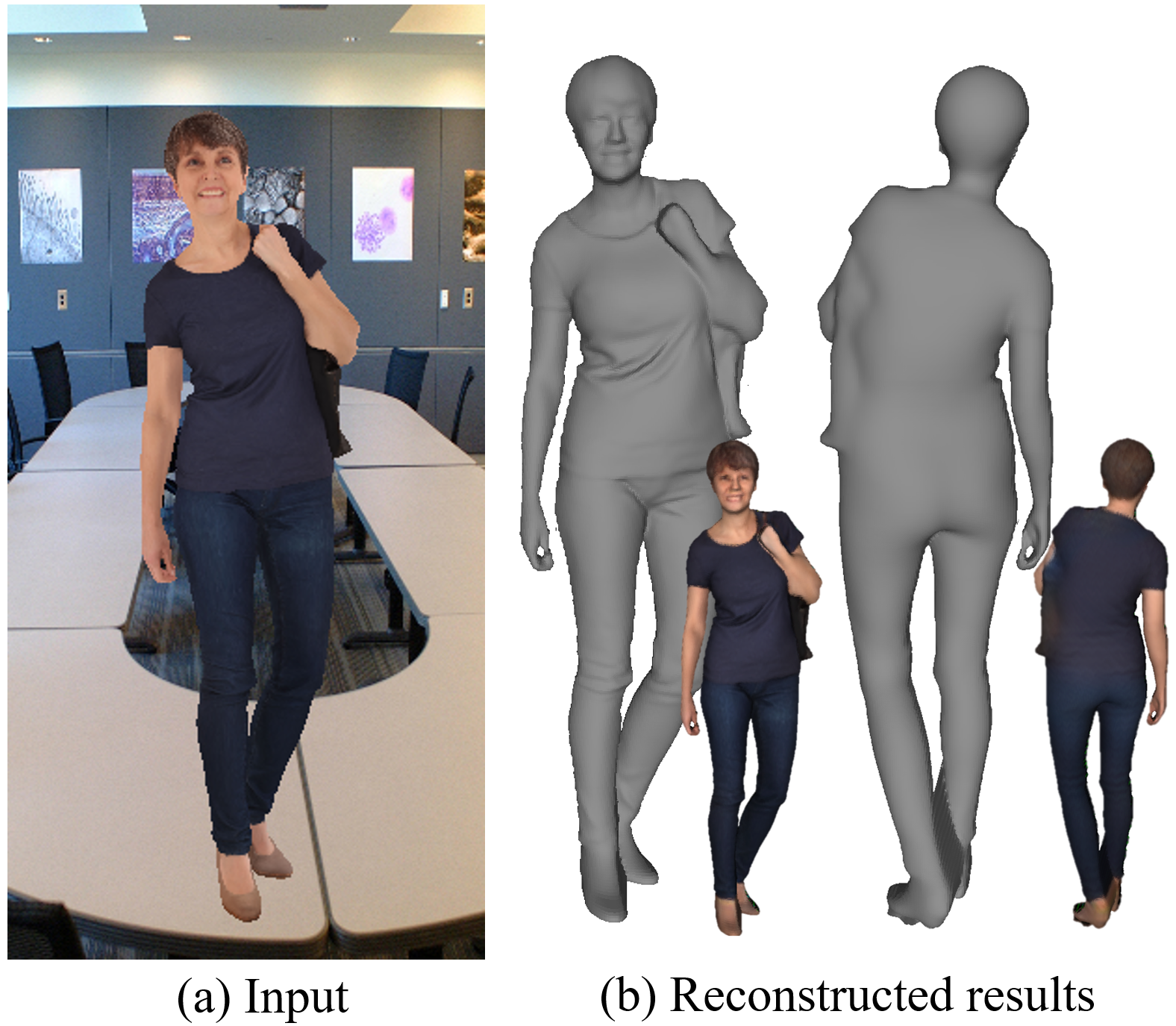}
\caption{An example of reconstructed human models. The predicted human model (b) is reconstructed by the proposed method from the input image (a).}
\label{fig:intro}
\end{figure}

Approaches to building virtual humans are diverse. Therefore, we categorize them into two groups: model-based and model-free approaches. The model-based approach first leverages a parametric 3D human model~\cite{SMPL_2015_Graph} for reconstruction~\cite{Zheng_2019_ICCV,Zheng_2021_PaMIR,Prokudin_2021_SMPLpix,Zhu_2021}. For example, the parametric 3D human model can used to construct initial implicit volumes~\cite{Zheng_2019_ICCV} or depth maps~\cite{Jafarian_2021_CVPR}. In general, this approach is robust under human pose variations, which is suitable for analyzing the movement of a human such as dancing. 

On the other hand, the model-free approach is good at digitizing dressed humans, including skirts or complex clothing, because it does not constrain the shape of the human model. Several studies tackle the human reconstruction problem by learning deep implicit functions~\cite{Varol_2018_ECCV,Saito_2019_ICCV,Saito_2020_CVPR,Mustafa_2021} or by predicting depth maps~\cite{Gabeur_2019_ICCV,Li_2019_CVPR,Jafarian_2021_CVPR,Tang_2019_ICCV,Tan_2020_self}, without employing structural constraints on the human model.
Notably, one of the breakthrough methods \cite{Saito_2019_ICCV} trains a deep implicit function to recover dressed human models from a single image. The deep implicit function is based on multi-layer perceptrons (MLPs) defined over the 3D space with the image-aligned features. Therefore, the implicit function can construct an implicit volume encapsulates the geometry of a 3D human model. One advantage of learning the implicit function is that it can build a large size volume by repeatedly calling the implicit function, \eg millions or billions of times. Ironically, this also increases the inference time cubically, which takes tens of seconds for high-resolution images.
Instead of training an implicit function, several studies train deep networks that predict the entire implicit volume~\cite{Varol_2018_ECCV,Jackson_2018_ECCV_Workshops} directly. However, this approach requires a large amount of memory, therefore, it is limited to taking low-resolution input compared to the approach mentioned above.
Another efficient approach to building a human model is to predict depth maps from an image. It is necessary to predict at least two depth maps to build a complete human model, \eg \cite{Gabeur_2019_ICCV} reconstructs a human model by merging predicted depth maps.

\begin{figure*}[t]
\centering
\includegraphics[width=2.1\columnwidth]{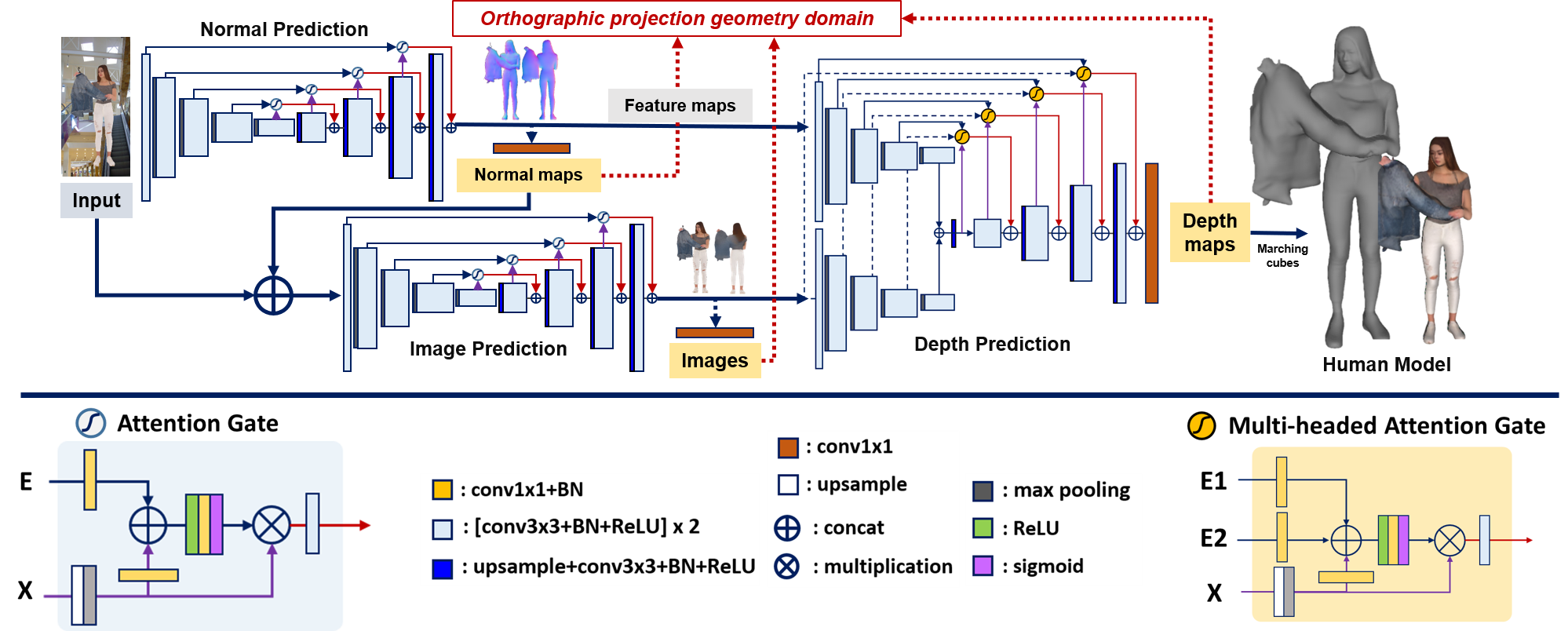}
\caption{An overview to our framework and key features. Our framework takes a single image as input, and predicts double-sided normal maps, shade removed images, and depth maps. The three networks are equipped with self-attention gates to extract meaningful information effectively. Moreover, we design a multi-headed attention network to take both photometric and geometric cues.}
\label{fig:framework}
\end{figure*}

Inspired by the previous works~\cite{Gabeur_2019_ICCV,Wang_2020_ECCV}, we propose a model-free approach to generate human models by predicting double-sided orthographic depth maps from a single image. The overall framework is illustrated in Fig.~\ref{fig:framework} and in Sec.~\ref{sec:proposed}. We summarize the contribution of our work as below. 
\begin{itemize}
\item We predict double-sided surface normals, shade-removed images, and depth maps that are seen from orthographic cameras placed at the front and backsides of a human, whereas the input is captured by a perspective camera. That is, our networks implicitly re-project the output orthographically, and we discuss its advantages in Sec.~\ref{sec:normal_and_image}. 
\item We propose a multi-headed Attention U-Net (mAUNet) to combine different modalities while selectively taking informative features through the multi-head attention gates. This scheme better preserves the details of humans, especially in facial regions and wrinkles in clothes.
\item We augment synthetic backgrounds during training to use our model for real-world applications, \ie images are captured in uncontrolled environments. Moreover, our method is robust under camera viewpoint variations benefited from orthographic depth map prediction.
\item We acquired a large number of human scans, \ie 18,831 scans, to train your network, which yields high-quality human model generation under pose and cloth variations. 
\end{itemize}

\section{Related Work}
\label{sec:related}

We review relevant studies focusing on a single image-based human reconstruction, which we categorize the existing approaches into three groups.

\noindent\textbf{Implicit volume prediction.}
Most studies predict implicit volumes to construct a complete 3D human model from a single image. \cite{Jackson_2018_ECCV_Workshops} reconstructs 3D human models by directly regressing the volume via a stacked hourglass network. BodyNet~\cite{Varol_2018_ECCV} extracts multiple reconstruction cues from an input image, \ie 2D pose and body segments, and merges them to predict the 3D human pose. After that, the volume prediction network takes the 3D human pose and the image to predict the implicit volume. Due to the memory demanding 3D convolution operations, the BodyNet takes 256$\times$256 resolution images as input, resulting in low-resolution human models. PIFu~\cite{Saito_2019_ICCV} trains a deep implicit function defined over a designated 3D space which constructs an implicit volume by repeatedly calling the implicit function. PIFuHD~\cite{Saito_2020_CVPR} extends PIFu to take high-resolution images, \ie 1024$\times$1024. It trains two implicit functions, one for constructing coarse resolution implicit volume and the other for enhancing details, extracted from double-sided high-resolution normal maps, translated from the input image. PIFuHD requires calling the implicit function 1024$^3$ + 512$^3$ times at inference time, for 1024$\times$1024 resolution input, limiting its usage, \eg, real-time applications. Geo-PIFu~\cite{He_2020_NeurIPS} suggests a more efficient way to build high-fidelity 3D models. They first predict a holistic implicit volume to preserve global 3D shape at a coarse level. In fine-level, they train details of an implicit function in a pixel-wise manner, similar to PIFuHD. \cite{Onizuka_2020_CVPR} propose the tetrahedral outer shell volumetric truncated signed distance function (TetraTSDF) to effectively predict human models, in which the part connection network regresses the detailed body shape within the tetrahedral volume. 
Note that one advantage of deep implicit volume prediction is that it can be extended to multi-view images~\cite{Yu_2021_function4d}, though we focus on a single image input in this paper.
%

\noindent\textbf{Implicit volume refinement.}
We also point out that the deep implicit volume can be refined through a deep network~\cite{Zheng_2019_ICCV,Zheng_2021_PaMIR}, where the initial volume can be constructed from various ways such as a parametric human model~\cite{Guler_2019_CVPR,SMPL_2015_Graph,SMPL-X_2019_Graph}, depth maps, or 3D body pose. \cite{Zheng_2019_ICCV,Zheng_2021_PaMIR} propose an image-guided volume-to-volume translation network. They use multi-scale volumetric feature transformation to fuse 2D image guidance information into a 3D volume. \cite{Mustafa_2021} suggests reconstructing multiple 3D human models from a single image, in which initially predicted volumes are refined through the hybrid feature decoder.

\noindent\textbf{Depth map prediction.}
Since it is unnecessary to employ 3D convolution layers for the depth prediction task, the depth map prediction approach can effectively build high-fidelity human models. Here, the depth map can be predicted for the front view~\cite{Jafarian_2021_CVPR,Tang_2019_ICCV,Tan_2020_self}, the back view~\cite{Wang_2020_ECCV}, both views, or more views~\cite{Gabeur_2019_ICCV,Jinka_2020_3DV}. \cite{Tang_2019_ICCV} proposes a single depth prediction network that exploits semantics and 3D joints, to leverage the normal map and initial depth prediction. Then, predicted results 
are merged and refined through recurrent convolution layers. Gabuer \etal \cite{Gabeur_2019_ICCV} predict double-sided depth maps, \ie visible and hidden surfaces, through a stacked hourglass network with adversarial training. The resultant depth maps are merged into a point cloud and meshed via Poisson reconstruction. Wang \etal propose NormalGAN~\cite{Wang_2020_ECCV}, which employs the discriminator network for the normal map instead of the depth map. Jinka \etal proposes propose PeelNet~\cite{Jinka_2020_3DV} to reconstruct a complete 3D shape through the peeled depth maps. Contrarily to conventional depth maps, it predicts multiple depth values per pixel. Therefore, it results in a more complete human model, especially in self-occluded regions. Jafarian and Park~\cite{Jafarian_2021_CVPR} propose an unsupervised approach to predict front depth maps. They utilize unlabelled social videos to train the network. It predicts initial depth by using DensePose~\cite{Guler_2018_DensePose} and refines the depth maps by minimizing geometric and photometric inconsistency between consecutive frames. Compared to the volumetric approach, the depth map prediction approach has not received much attention. We hope our work triggers new studies on this topic.

\section{Proposed Method}
\label{sec:proposed}

Our framework consists of three networks as shown in Fig.~\ref{fig:framework}. The first network predicts surface normal maps from an input image, the second network predicts shade-removed images, and the last network predicts depth maps. Note that all the outputs are double-sided, \ie front and back views, and re-projected orthographically, whereas the input is captured by a perspective camera.

\subsection{Normal and image prediction networks}
\label{sec:normal_and_image}
Initially, the normal prediction network extracts the surface normals from an input image, 
\begin{equation}\label{eq:network_normal}
  \mathbf{\bar{N}} = \mathcal{N}(I_\text{input}),
\end{equation}
where the network $\mathcal{N}(\cdot)$ predicts double-sided normal maps, $\mathbf{\bar{N}}={\{N_f, N_b\}}$, where subscripts ${f}$ and ${b}$ indicate front and back views. Here, our empirical finding is that predicting depth maps from initially predicted normal maps better preserves the details of a human body, than predicting depth maps directly. Afterward, we predict shade-removed images from the predicted normal maps and the input image,
\begin{equation}\label{eq:network_normal}
  \mathbf{\bar{I}} = \mathcal{C}({I}_\text{input}\oplus\mathbf{\bar{N}}),
\end{equation}
where $\mathbf{\bar{I}}$ indicates the shade removed images, \ie $\mathbf{\bar{I}}=\{I_f, I_b\}$.
We employ the predicted normal maps to eliminate shadings from the input image to avoid unrealistic rendering in a virtual space such as a metaverse space.
Note that we focus on generating visually plausible images rather than predicting intrinsic components of images, which requires to predict lighting sources and material properties in addition to the surface normals.
Through the two networks, we extract geometric and photometric cues from the input image.
Here, we employ Attention UNet (AUNet)~\cite{Oktay_2018_atunet}, for normal and image prediction networks. 

\vspace{3pt}\noindent\textbf{Orthographic re-projection.}
The most important feature of our framework is that the normal and image prediction networks implicitly re-project outputs to orthographic views, and the depth prediction network naturally gives orthographic depth maps as well.
Moreover, it is worth noting the advantages of the orthographic re-projection scheme.
First, the orthographic projection captures more regions of a human body than perspective projection, as also stated in~\cite{Wang_2020_ECCV}. In more general words, when the same number of depth maps are used to build a 3D human model, the completeness of the model is higher for the orthographically projected depth maps.
Second, the front and back views are perfectly aligned. Therefore, it is more suitable for predicting double-sided normal maps, depth maps, and images.
In the case of aligned orthographic depth maps, they can be easily integrated into a single implicit volume since the implicit volume can be constructed identically for the two depth maps.
%
%
Third, if we add more views to be predicted, $y$ coordinates across different views can be easily aligned, making the human modeling problem easier and simpler.
In summary, it is a simple idea to predict implicitly re-projecting outputs, and we claim that this idea helps digitizing humans. 

\subsection{Depth prediction network}
Given normal maps and color images, we predict the depth maps through a multi-headed Attention UNet (mAUNet), 
\begin{equation}\label{eq:network_depth}
  \mathbf{\bar{D}} = \mathcal{D}(\phi(\mathbf{\bar{I}}), \phi(\mathbf{\bar{N}})), 
\end{equation}
where $\mathbf{\bar{D}}=\{D_f, D_b\}$, indicates the double-sided depth maps.
Here, instead of feeding normal maps and images directly to the network, we feed feature maps $\phi(\mathbf{\bar{I}})$ and $\phi(\mathbf{\bar{N}})$ extracted at the last layer of each network, \ie the feature maps are converted to the normal maps and the images through the $1\times1$ output layers, as illustrated in Fig.~\ref{fig:framework}.
This is not to lose meaningful information from the feature maps. 
Therefore, the depth prediction network takes geometric and photometric feature maps through the multi-head encoders, combines them in the latent space, and predicts accurate depth maps with the aid of multi-headed attention gates (mAG). The mAG extracts and combines information from different modalities with multi-headed encoder networks.

\subsection{Loss functions}
\label{sec:loss}
We exploit L1 loss, perceptual loss, and structural similarity index measure (SSIM) loss functions that are popularly used in the literature, and we explain them in Eqs.~\eqref{eq:l1_loss}-\eqref{eq:ssim_loss}.
Overall, we minimize the following loss function to train the entire framework in an end-to-end fashion, 
\begin{equation}\label{eq:total_loss}
  \mathcal{L} = \mathcal{L}_{N} + \mathcal{L}_{C} + \mathcal{L}_{D}, 
\end{equation}
where each term is defined for one of the three networks. 
First, $\mathcal{L}_{N}$ is the loss function for the normal prediction network, 

\begin{equation}
\label{eq:loss_normal}
  \mathcal{L}_N =\lambda_1{L}_\text{L1}\mathbf{\bar{N}} + {\lambda}_{2}{L}_\text{SSIM}\mathbf{\bar{N}},
\end{equation}

where the first term minimizes the L1 loss and the second term minimizes structural dissimilarity between predicted and ground truth normals. 
Second, $\mathcal{L}_{C}$ is designed for the color prediction network, 
\begin{equation}\label{eq:loss_normal2color}
  \mathcal{L}_{C} = \lambda_{3}\mathcal{L}_{\text{L1}}(\mathbf{\bar{I}}) + {\lambda}_{4}\mathcal{L}_\text{PER}(\mathbf{\bar{I}}). 
\end{equation}
Similarly, the first term reduces the L1 loss and the second term minimizes perceptual loss between the predicted and ground truth images. 
Lastly, $\mathcal{L}_{\mathcal{D}}$ is used to train the depth prediction network,
\begin{equation}\label{eq:loss_normal2depth}
  \mathcal{L}_{\mathcal{D}} = \lambda_{5}{L}_\text{L1}(\mathbf{\bar{D}}) + \lambda_{6}{L}_\text{SSIM}(\mathbf{\bar{D}}) + 
  \lambda_{7}{L}_\text{L1}(\nabla\mathbf{\bar{D}}) + \lambda_{8}{L}_\text{SSIM}(\nabla\mathbf{\bar{D}}),
\end{equation}
which reduces L1 and SSIM losses for both depth maps and their gradients, to better preserve the details of human body. In addition, the L1 loss ${L}_\text{L1}$ is formulated by,
%

\begin{table*}[t]
\setlength{\tabcolsep}{3pt}
\footnotesize
\centering
\caption{Single-view human reconstruction evaluation results on the RenderPeople, BUFF, and our dataset. The table is divided into two main blocks where the numbers in the upper block are evaluated with moderate test sets and the lower block is evaluated with difficult cases. The errors are calculated by using the pre-trained models from the authors and the best results are shown in blue.}
\label{table:eval1}
\begin{center}
\vspace{-10.0pt}
\renewcommand{\arraystretch}{1.2} 
	\begin{tabular}{c | c | c | c | c | c | c | c | c | c }
		\noalign{\hrule height 1pt}   
		  & \multicolumn{3}{c|}{RenderPeople} & \multicolumn{3}{c|}{BUFF} & \multicolumn{3}{c}{Custom} \\ \cline{2-10}
		 & {P2S$\downarrow$} & {Chamfer$\downarrow$} & {Normal$\downarrow$} & {P2S$\downarrow$} & {Chamfer$\downarrow$} & {Normal$\downarrow$} & {P2S$\downarrow$} & {Chamfer$\downarrow$} & {Normal$\downarrow$} \\ 
		 \hline
		 PIFu(M) & {1.26} & {1.18} & 0.1377 & 1.55 & 1.49 & 0.1612 & 1.49 & 1.52 & 0.1480\\
		 PIFuHD(M) & \bblue{1.10} & {1.08} & 0.1562 & \bblue{0.86} & 0.83 & 0.1213 & \bblue{1.38} & 1.35 & 0.1661 \\
		 PaMIR(M) & 1.57 & 1.57 & 0.1489 & 1.45 & 1.43 & 0.1433  & 2.06 & 1.87  &  0.1528  \\
		 Ours(M) & {1.32} & \bblue{1.01} & \bblue{0.0960} & 1.16 & \bblue{0.80} & \bblue{0.1171} & 1.56 & \bblue{1.21} & \bblue{0.1084} \\
		 \hline\hline
		 PIFu(H) & {2.56} & {2.67} & 0.1682 & 2.29 & 2.16 & 0.1923 & 2.30 & 2.67 & 0.1846 \\
		 PIFuHD(H) & {2.15} & {2.20} & 0.2026 & 1.50 & 1.43 & 0.1479 & 2.10 & 2.17 & 0.2218 \\
		 PaMIR(H) & 2.25 & 2.37 & 0.1924 & 1.95 & 1.94 &  0.1703  & \bblue{2.05}  &  2.09 &  0.1880  \\
		 Ours(H) & \bblue{1.88} & \bblue{1.69} & \bblue{0.1624} & \bblue{1.45} & \bblue{1.10} & \bblue{0.1329} & {2.06} & \bblue{1.75} & \bblue{0.1748}\\ 
         \hline
         \noalign{\hrule height 1pt} 
	\end{tabular}
\end{center}
\vspace{-5pt}
\end{table*}

\begin{figure*}[t]
\centering
\vspace{-10.0pt}
\includegraphics[width=1.7\columnwidth]{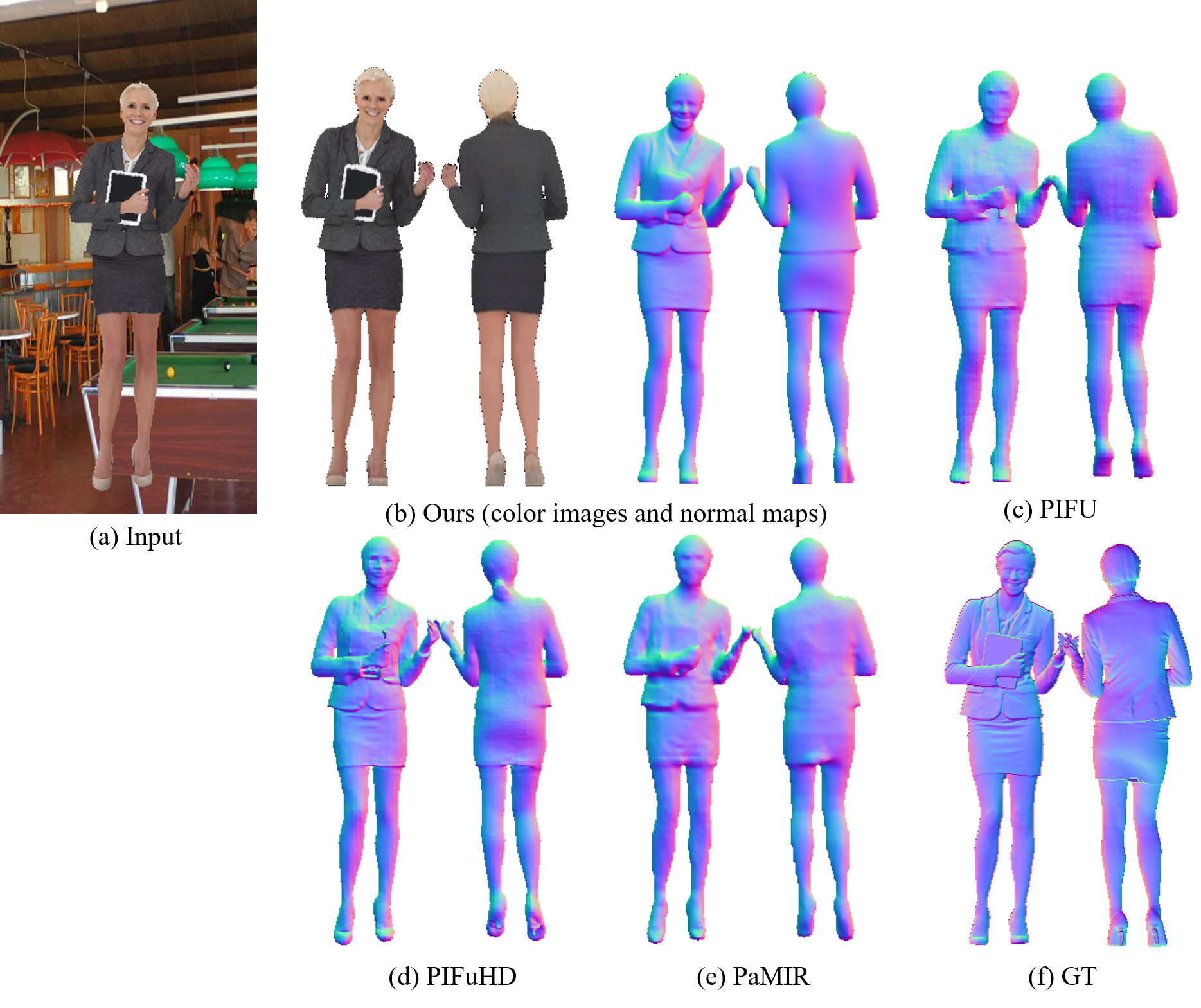}
\caption{Qualitative comparisons of reconstructed results. Our method shows accurate results compared to (c)-(e); (c) cannot recover details of human body whereas normal maps from (d) and (e) are frequently misaligned with ground truth models.}
\label{fig:qual_eval}
\end{figure*} 

\begin{figure*}[t]
\centering
\vspace{-10.0pt}
\includegraphics[width=1.7\columnwidth]{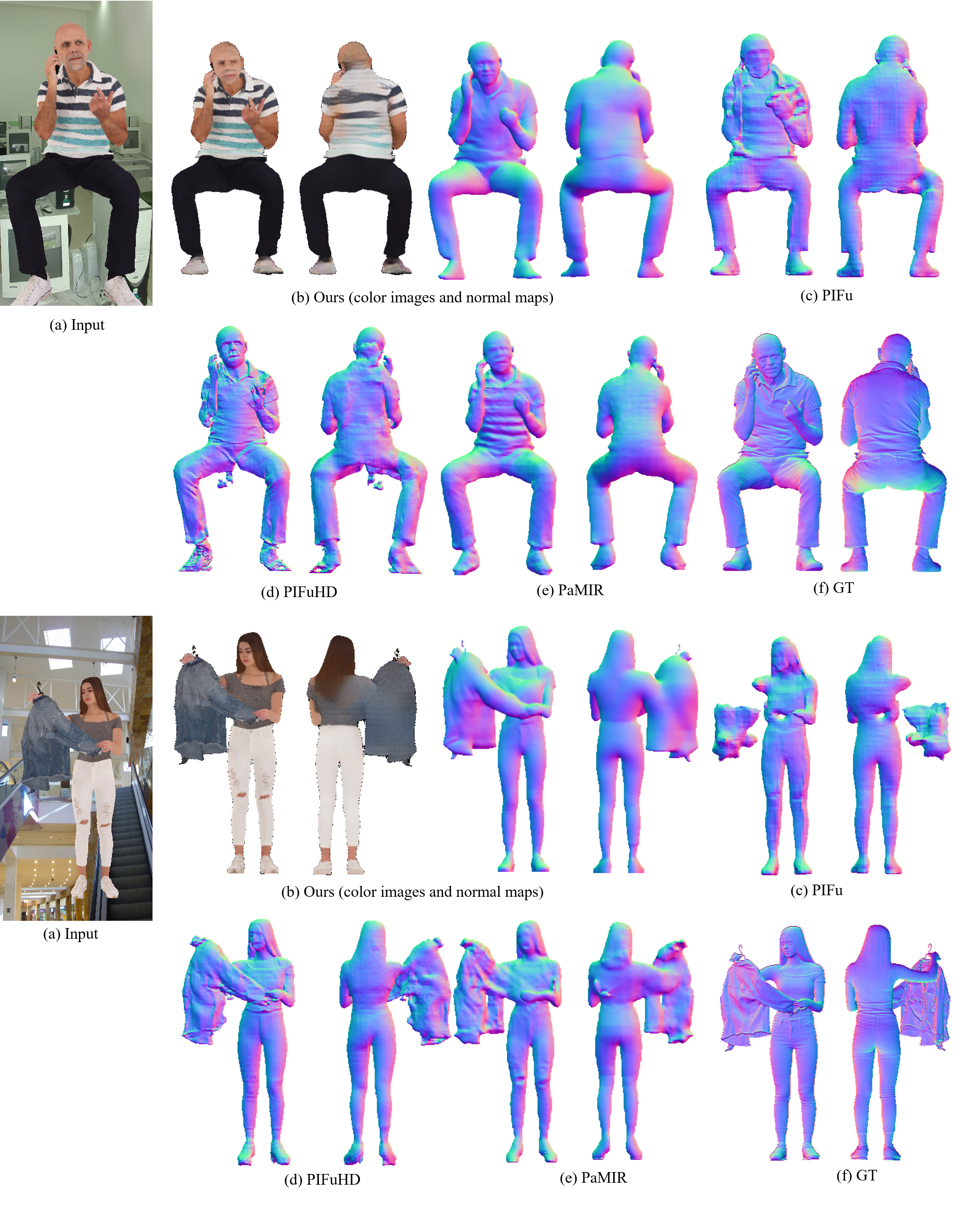}
\caption{Qualitative comparisons of reconstructed results. Our method shows accurate results compared to (c)-(e); (c) cannot recover details of human body whereas normal maps from (d) and (e) are frequently misaligned with ground truth models.}
\label{fig:qual_eval}
\end{figure*} 

\begin{equation}\label{eq:l1_loss}
  {L}_{L1}(\mathbf{\bar{x}}) = \dfrac{1}{N}\underset{\forall p}{\Sigma} || {\bar{x}}_p - {x^*}_p||_{1},
\end{equation}
\noindent
where $\bar{x}_p$ is the pixel of predicted output and $x^*_p$ is its corresponding label. $\mathbf{\bar{x}}$ can be either predicted normal map, depth map, and image, depending on the network. $N$ is the number of pixels.
It is worth noting that we put zero to normal maps, images, and depth maps if the pixel does not belong to the human. 
Therefore, our framework implicitly extract foreground objects, \ie humans, in uncontrolled environments without employing separate segmentation network or losses.
${L}_\text{PER}$ stands for the perceptual loss between images which has frequently used in the literature~\cite{Ledig_SRGAN16}. It minimizes the loss between the images,
\begin{equation}\label{eq:perceptual_loss}
        {L}_\mathrm{PER}(\mathbf{\bar{I}})=
        ||\mathcal{G}_{\phi_{l}}(\mathbf{\bar{I}})-\mathcal{G}_{\phi_{l}}(\mathbf{I}^*)||_{1}, 
\end{equation}
where $\mathbf{I}^*$ is the ground truth image, $\phi_{l}$ is the $l$th latent space feature predicted by the VGG16 network, and $\mathcal{G}_{\phi_{l}}(\cdot)$ calculates a Gram matrix by using $\phi_{l}$.
${L}_\text{SSIM}$ is the SSIM loss, 
\begin{equation}\label{eq:ssim_loss}
  {L}_{\text{SSIM}}(\mathbf{\bar{x}}) = 1 - \dfrac{1}{N}\underset{\forall p}{\Sigma}\dfrac{(2{\mu}_{\bar{x}}\mu_{x^*}+c_1)(2{\sigma}_{\bar{x}x^*} + c_2)}{({\mu}_{\bar{x}}^2 + \mu_{x^*}^2 + c_1)({\sigma}_{\bar{x}}^2 + \sigma_{x^*}^2 + c_2)},
\end{equation}
where mean and variance values, $\mu$ and $\sigma$ are calculated within an 5$\times$5 size window centered at $p$. $c_1$ and $c_2$ are set to $0.01^2$ and $0.03^2$, respectively. Moreover, the input is assumed to be normalized in the range $(0, 1)$.

\section{Experimental Results}
\label{sec:exp}

\subsection{Human model dataset}
\label{sec:dataset}
We acquired human scan models from our scan booth that consists of 80 DSLR cameras, and additionally purchased 106 high-quality human models from RenderPeople~\footnote{https://renderpeople.com/}, and downloaded 176 human models from BUFF~\cite{Zhang_2017_BUFF}. As a result, our dataset consists of human models of 18,549 people, ages from 10 to 60's, from multiple countries but mostly Asians.
Note that this is the largest amount of training data ever used before, and we believe that the success of human modeling comes with high-quality, large-scale training data. Therefore, we will open our data partially, which are allowed to be opened publicly.

\noindent\textbf{Training data generation.}
To train our network, we generated semi-synthetic images as follows. 
We shifted human models to the origin, \ie the centroid of each model, so we can assume the human is always placed at the center of the image. 
Second, for a shifted model, we rotated the model horizontally from -40 to 40 with an interval of 10 degrees.
Third, with rotated human models, we render images \wrt the virtual camera in front of the model, \ie at [0, 0, -1], looking towards the model. The camera's field of view (FoV) is 50 degrees, and the resolution of images is 512$\times$256.
We generated front and back normal maps, depth maps, and shade removed images using the orthographic projection, where we also put 180 virtual diffuse lighting sources around the model having random colors.
In summary, one model was rotated eight times; for each rotated position, we rendered double-sided depth maps and shade removed images. Then, normal maps were generated from the depth maps. If we regard pairs of depth maps, images, and normal maps as one sample, 166,941(162,003 for training/4,938 for validation) samples are generated from our dataset, 870(860/10) samples are generated from the render people dataset, and 1,584(1,564/20) samples are generated from the BUFF dataset. 
We generated an input image for each sample after superimposing the synthetic background behind the model.
Background images were randomly selected from the indoor scene dataset~\cite{Quattoni_2019_indoor}, sample images are shown in Fig.~\ref{fig:qual_eval}.

\noindent\textbf{Data augmentation and parameter settings.}
Before feeding input image to the network, we normalized color images with VGG parameters, \ie mean colors [0.485, 0.456, 0.406] with standard deviation of [0.229, 0.224, 0.225]. Samples are shown in Fig.~\ref{fig:qual_eval}. We set $\lambda_1$-$\lambda_8$ to [0.9, 0.1, 0.85, 0.15, 0.45, 0.05, 0.45, 0.05], respectively. We trained the entire networks for 50 epochs, with all datasets without fine-tuning for each dataset. The learning rate was set to 1e-4, and decay by the factor of 0.95 for every epoch. We used Adam optimizer and LambdaLR scheduler. The training process was carried out with two RTX A6000 GPUs, which took 200 hours to train for 50 epochs. The batch size was set to 10.

\begin{figure}[t]
\centering
\vspace{-10.0pt}
\includegraphics[width=1.\columnwidth]{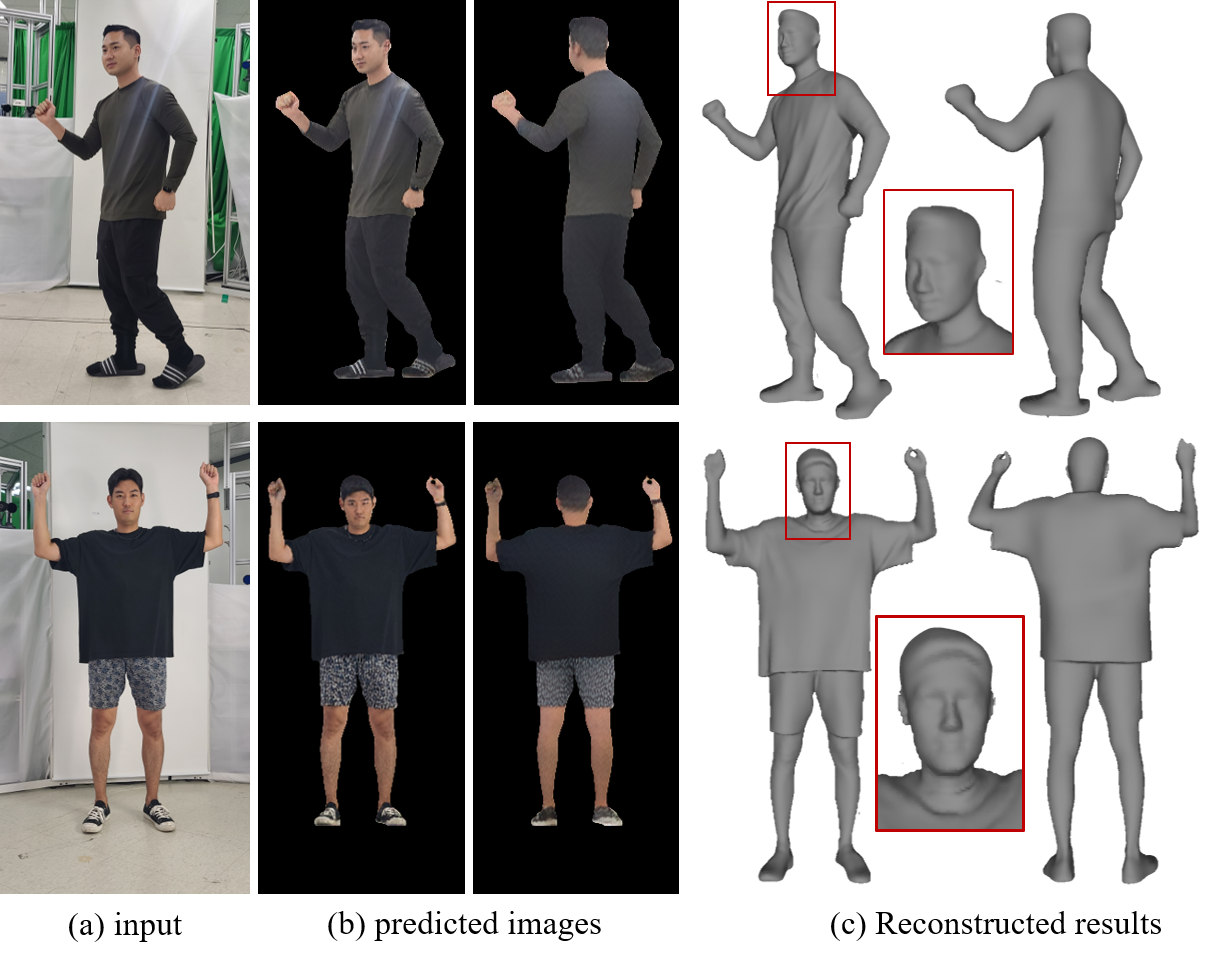}
\caption{Example results obtained by images, captured by a smart phone.}
\label{fig:qual_general}
\end{figure} 

\subsection{Quantitative and qualitative evaluation}

\noindent\textbf{Evaluation metrics.}
For evaluation, we use popularly used metrics in the literature: average point-to-surface distance (P2S), Chamfer distance, and average surface normal error. Metric units are centimeters for P2S and Chamfer, and radian for surface normal.

\noindent\textbf{Quantitative comparison.}
We compare our method with the state-of-the-art (SOTA) methods, including PIFu~\cite{Saito_2019_ICCV}, PIFuHD~\cite{Saito_2020_CVPR}, and PaMIR~\cite{Zheng_2021_PaMIR} using human scan data from RenderPeople, BUFF~\cite{Zhang_2017_BUFF}, and our dataset.
We selected 20 moderate and 20 hard test cases for RenderPeople and BUFF datasets and 25 moderate and 25 hard cases for our dataset to see the characteristics of different approaches depending on the difficulties of data. 
Table~\ref{table:eval1} compares the errors for all datasets; the proposed method outperforms existing methods in most cases. Furthermore, our method shows significantly better results in terms of average normal errors, which confirms that our method is good at recovering the details of a human body.
Although the proposed method does not guarantee complete human reconstruction, which fails at self-occluded regions, it performs better than volume prediction approaches, especially when the degree of reconstruction difficulties increases.
During inference, our method runs at 5 fps with a single RTX 2080Ti excluding mesh generation, about 70 times faster than PIFuHD.

\noindent\textbf{Qualitative comparison.}
Figure~\ref{fig:qual_eval} visually compares the results of different approaches. Here, the normal maps are generated from the reconstructed human models by projecting the models orthographically to see the quality of models and the alignment between the predicted and ground truth models.
The results show that our method and PIFu align appropriately with the ground truth normal maps, whereas the other methods show misaligned results in foot regions. This is the major source of quality degradation for PIFuHD and PaMIR. 
%
%

\begin{table}[t]
\setlength{\tabcolsep}{5pt}
\footnotesize
\centering
\caption{Ablation study. The first row indicates that we predict depth maps directly from the image, the second row replaced attention gates with skip connections, and the last indicates the full model. }
\label{table:eval1}
\begin{center}
\vspace{-5.0pt}
\renewcommand{\arraystretch}{1.2} 
	\begin{tabular}{c | c | c | c | c | c | c }
		\noalign{\hrule height 1pt}   
		  & \multicolumn{2}{c|}{P2S$\downarrow$} & \multicolumn{2}{c|}{Chamfer$\downarrow$} & \multicolumn{2}{c}{Normal$\downarrow$} \\ \cline{2-7}
		 & M & H & M & H & M & H \\
		 \hline
		 Direct Depth & {1.46} & {1.99} & 1.16 & 1.73 & 0.1104 & 0.1759 \\
		 w/o Attention & {1.38} & {2.03} & 1.07 & {1.81} & 0.1054 & 0.1810 \\
		 Full & {1.32} & {1.88} & 1.01 & 1.69 & 0.0960 & 0.1624 \\
         \hline
         \noalign{\hrule height 1pt} 
	\end{tabular}
\end{center}
\end{table}

\begin{figure}[t]
\centering
\includegraphics[width=1.0\columnwidth]{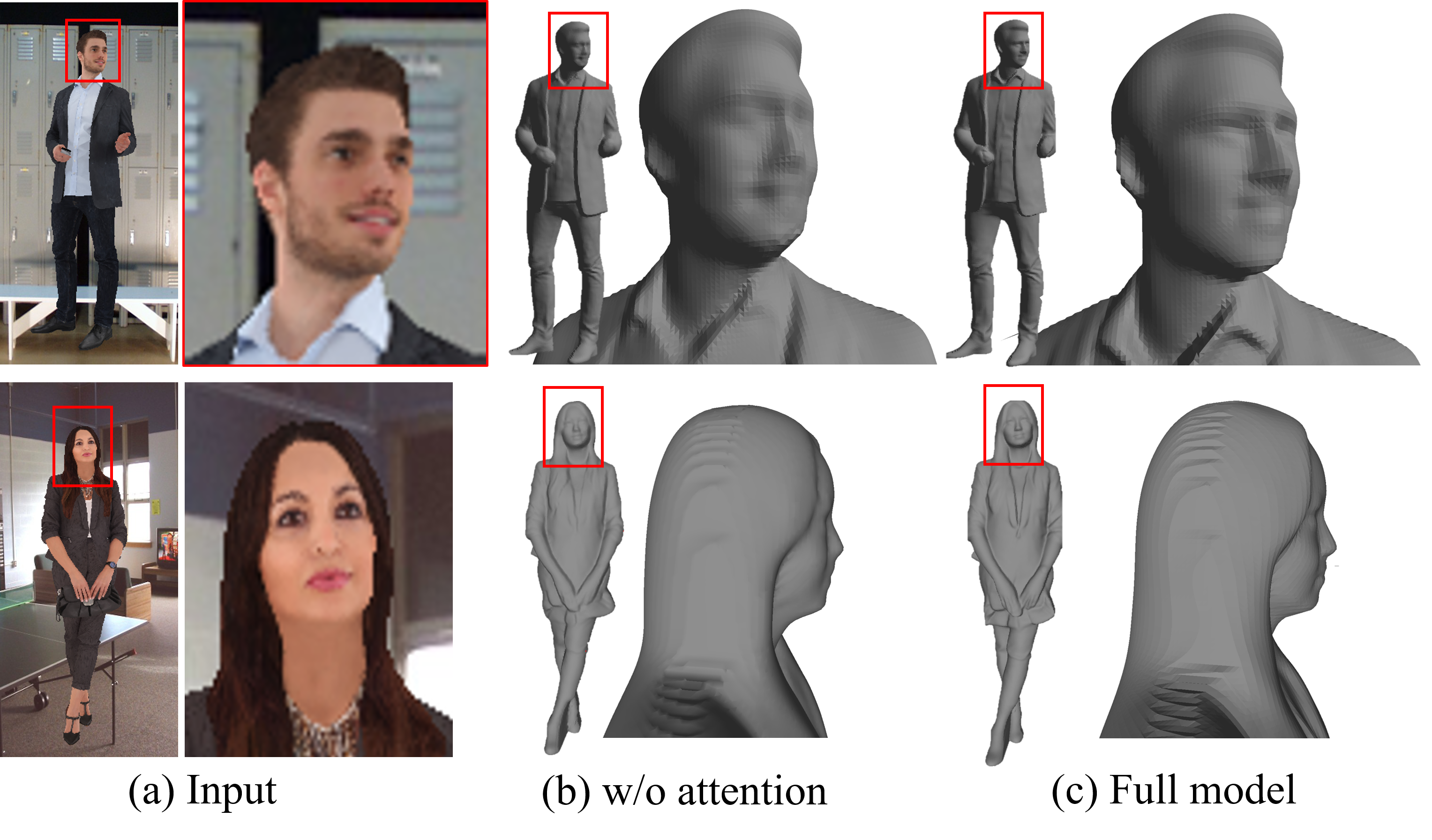}\vspace{-10pt}
\caption{Qualitative comparison for different network architectures.} 
\vspace{-10pt}
\label{fig:ablation_result}
\end{figure}

\noindent\textbf{Generalization.} We also apply our method to various input sources, an animation character and a real image, \ie, captured by a smartphone. As shown in Fig.~\ref{fig:qual_general}, the results are slightly degraded compared to the test dataset because the animation character has different characteristics compared to the real image. In case of the real image, we confirm that our method works properly as long as the background is simple and a single human is standing in the middle. 

\subsection{Ablation study}
\label{sec:ablation}
We evaluated our framework while differentiating network structures to see the necessity of key ideas.
Table~\ref{table:eval1} shows evaluation results under different configurations using RenderPeople dataset. First, we directly predicted depth maps instead of predicting normal maps with the same network, \ie ATUNet. Secondly, we replaced all the attention gates with the conventional skip connections to see the impact of the self-attention mechanism. 
As described in the table, normal prediction and attention gates play an important role in building 3D human models from a single image. Moreover, the direct depth prediction itself shows superior performance than existing methods. We believe this is because we used significantly larger training data than existing methods. 

\subsection{Limitations}
\begin{figure}[h!]
\centering
\includegraphics[width=0.9\columnwidth]{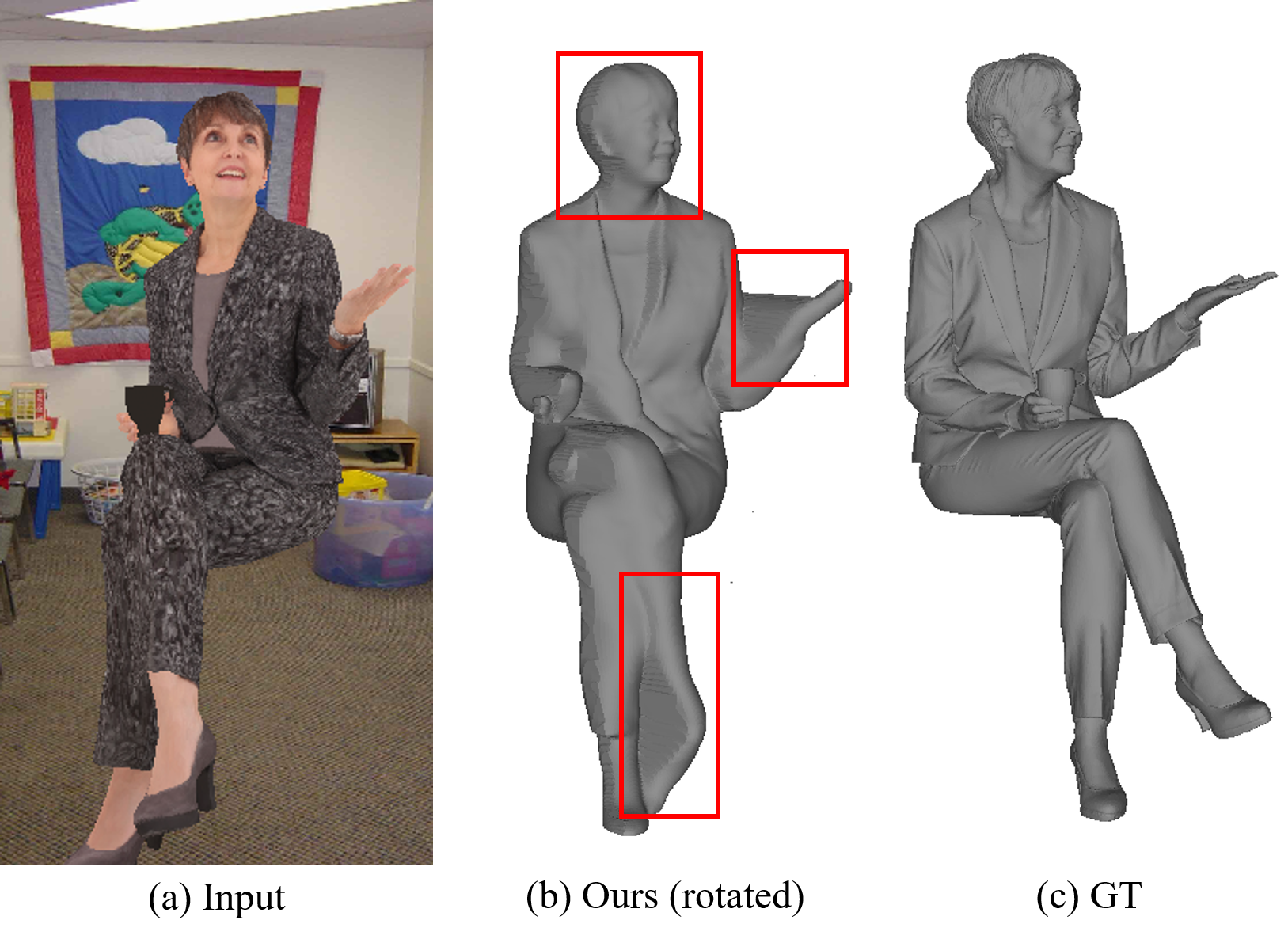}
\vspace{-5pt}
\caption{Highlighted regions show the limitations of the current approach.} 
\vspace{-10pt}
\label{fig:failure_case}
\end{figure}

From the perspective of ray-casting, double-sided depth maps can only reconstruct the first and last met surfaces. Therefore, its usage is limited to humans with a group of simple postures such as the A or T pose. 
In other words, our method fills the self-occluded regions as shown in Fig.~\ref{fig:failure_case} where we rotated the model. 
To remedy this limitation,  we need to predict additional views, \eg left and right views, or to multiple depth values as in~\cite{Jinka_2020_3DV}. 
The second problem we observed is the dataset bias. Since our dataset mainly consists of Asians, it slightly softens the shape of faces and noses. 
These limitations cast new problems in the depth map prediction and data-driven reconstruction approaches.

\section{Conclusion}
\label{sec:conclusion}

We have proposed a framework for digitizing humans by predicting double-sided depth maps from a single image. The key to our framework is two-fold. The first is to re-project outputs \wrt the orthographic cameras. The second is a multi-headed attention network to take the informative features from different modalities selectively. Our networks are trained with the huge amount of human scan models, and the experimental results confirmed the single image-based human digitization is now applicable for various applications. We hope our study lays a stepping stone between the real world and the meta-universe.

{\small
\bibliographystyle{ieee}
\bibliography{main.bib}
}

\end{document}